\documentclass{article}

\usepackage[accepted]{icml2023}

\usepackage[autonum,blackhypersetup]{shortex}
\usepackage{siunitx}
\usepackage[mode=tex]{standalone}
\usepackage{tikz}
\newcommand{\tikzmark}[1]{\tikz[overlay,remember picture] \node (#1) {};}
\usetikzlibrary{positioning, calc, intersections}

\usepackage[capitalize,noabbrev]{cleveref}

\usepackage[textsize=tiny]{todonotes}

\newcommand{\dd}{\mathop{}\!\mathrm{d}}

\begin{document}

\twocolumn[
\icmltitle{Uncertain Evidence in Probabilistic Models and Stochastic Simulators}

\begin{icmlauthorlist}
\icmlauthor{Andreas Munk}{ubc}
\icmlauthor{Alexander Mead}{ubc}
\icmlauthor{Frank Wood}{ubc,iai,mila}
\end{icmlauthorlist}

\icmlaffiliation{ubc}{Department of Computer Science, University of British Columbia, Vancouver, B.C., Canada}
\icmlaffiliation{iai}{Inverted AI Ltd., Vancouver, B.C., Canada}
\icmlaffiliation{mila}{Mila, CIFAR AI Chair}

\icmlcorrespondingauthor{Andreas Munk}{amunk@cs.ubc.ca}

\icmlkeywords{Machine Learning, Inference, Probabilistic Programming}

\vskip 0.3in
]

\printAffiliationsAndNotice{}  

\begin{abstract} 
    We consider the problem of performing Bayesian inference in probabilistic
    models where observations are accompanied by uncertainty, referred to as
    ``uncertain evidence.'' We explore how to interpret uncertain evidence, and
    by extension the importance of proper interpretation as it pertains to
    inference about latent variables. We consider a recently-proposed method
    ``distributional evidence'' as well as revisit two older methods: Jeffrey's
    rule and virtual evidence. We devise guidelines on how to account for
    uncertain evidence and we provide new insights, particularly regarding
    consistency. To showcase the impact of different interpretations of the same
    uncertain evidence, we carry out experiments in which one interpretation is
    defined as ``correct.'' We then compare inference results from each
    different interpretation illustrating the importance of careful
    consideration of uncertain evidence.
\end{abstract}

\section{Introduction}\label{sec:introduction}

In classical Bayesian inference, the task is to infer the posterior distribution
$p(x|y) \propto  p(y,x)$ over the latent variable $x$ given (an observed) $y$.
The joint distribution (or model), $p(y,x)$, is assumed known, and is typically
factorized as $p(y,x)=p(y|x)p(x)$ where $p(y|x)$ and $p(x)$ is the likelihood
and prior respectively. This paper deals with the case where $y$ is not observed
exactly; rather it is associated with uncertainty\footnote{Ideally one would
remodel the system to account for such uncertianties, but this is rarely easy to
do.} which we refer to as ``uncertain evidence.'' This is a fairly common
scenario as these uncertainties may stem from: observational errors; distrust in
the source provding $y$; or when $y$ is derived (stochastically) from some other
data.

\begin{table}
  \caption{Uncertain observation of the time $t$ in the ball dropping example.}
  \label{tab:times-uncertain}
  \begin{center}
  \begin{small}
  \begin{sc}
  \begin{tabular}{ccc}
        \toprule 
        & Value $[\si{\second}]$ & $\pm[\si{\second}]$ \\
        \midrule
        $t$ & 0.5 & $0.05$ \\
        \bottomrule
  \end{tabular}
  \end{sc}
  \end{small}
  \end{center}
  \vskip -0.1in
\end{table}

As a running example, consider the experiment of recording the time $t$ it takes
for a ball to drop to the ground in order to determine the acceleration due to
gravity, $g$. Taking some prior belief about the value of $g$, we may solve this
problem using Bayesian inference. That is, we infer $p(g|t)\propto p(g)p(t|g)$,
where $p(g)$ is the prior density of $g$ and $p(t|g)$ is the likelihood
representing the physical model (or simulation) of the time $t$ given $g$. In
this setup, the uncertainty about $t$ given $g$ would be due to neglecting air
resistance or ignoring variations in the distance the ball drops as a result of
vibrations etc. Assume next, that the observations (or data) is given as in
\cref{tab:times-uncertain}. It is not immediately obvious how the uncertainty
relates to $y$ and there are arguably at least two valid interpretations of the
information in \cref{tab:times-uncertain}: (1) it describes a distribution of
the real time $t$. For example, the real time is normally distributed with mean
$0.5\si{\second}$ and standard deviation $0.05\si{\second}$. (2) It describes
additional uncertainty on the predicted time and the observed value is, indeed,
$0.5\si{\second}$. For example, given the predicted time $t$ the observed time
$\hat{t}$ is normally distributed with mean $t$ and standard deviation
$0.5\si{\second}$. Importantly, in either case the uncertainty can be
represented with a given \textit{external}\footnote{In this context, external
refers to a distribution provided from some external source.} distribution,
$q(\cdot|\cdot)$, which describes a stochastic relationship between $t$ and an
auxiliary variables $\zeta$. We consider in case (1) and (2) the distributions
$q(t|\zeta)$ and $q(\zeta|t)$. In the former case $\zeta$ is left implicit
(something gave rise to the uncertainty), and in the latter $\zeta=\hat{t}$ and
the observation is $\hat{t}=0.5\si{\second}$. These two approaches are
fundamentally different operations that may lead to profoundly different
inference results.

The topic of observations associated with uncertainty has been studied since at
least 1965~\citep{jeffrey1965logic}. Of particular relevance are the work
of~\citet{jeffrey1965logic} and~\citet{shafer1981jeffrey};
and~\citet{pearl1988probabilistic}, giving rise to \textit{Jeffrey's
rule}~\citep{jeffrey1965logic, shafer1981jeffrey} and \textit{virtual
evidence}~\citep{pearl1988probabilistic}. In the example above, inference using
approach (1) and (2) corresponds to Jeffrey's rule and virtual evidence
respectively. Since then other approaches, closely related to Jeffrey's rule and
virtual evidence has been
proposed~\citep[e.g.][]{valtorta2002soft,tolpin2021probabilistic,doi:10.1080/03610926.2020.1838545}.
While each approach has its own merits and is applicable under (almost) the same
circumstances, the original literature and most prior work comparing these
methods,
\citep[e.g.,][]{pearl2001two,valtorta2002soft,chan2005revision,benmrad2013understanding,tolpin2021probabilistic},
are reluctant to take a concrete stand on when each is more appropriate. 

This paints an obfuscated picture of what to do, practically, when presented
with uncertain evidence. This obfuscation becomes problematic when practitioners
outside the field of statistics deal with uncertain evidence and look to the
literature for ways to address it, especially, considering the increased use of
Bayesian inference in high-fidelity simulators and probabilistic models
\citep[e.g.,][]{pmlr-v89-papamakarios19a,baydin2019efficient,lavin2021simulation,liang2021kepler90,vandeschoot2021bayesian,wood2022planning,PhysRevD.105.063017,munk2022probabilistic}.
For examples, in physics it is not uncommon that likelihoods are given
relatively ad-hoc forms where some notion of ``measurement error'' is attached
to uncertain observations. However, the underlying (stochastic) \emph{physical}
model is usually taken to be understood perfectly. For example when inferring;
the Hubble parameter via supernovae brightness \citep[e.g.,][]{Riess2022};
pre-merger parameters of black-hole/neutron star binaries via gravitational
waves \citep[e.g., ][]{Thrane2019, Dax2021}; neutron star
orbital/spin-down/post-Newtonian parameters via pulsar timings
\citep[e.g.,][]{Lentati2014, Vigeland2014}; planetary orbital parameters via
radial velocity/transit-time observations \citep[e.g.,][]{Schulze-Hartung2012,
Feroz2014, liang2021kepler90}. In most cases a Gaussian likelihood is assumed
for the data, but exactly how the error relates to the data generation process
is not specified. If uncertainties about simulator/model observations arise
given external data, then usually Jeffrey's rule would apply, but it appears
that virtual evidence is more often employed.

It is the purpose of this paper to provide novel insights, theoretical
contributions and concrete guidance as to how to deal with observations with
associated uncertainty as it pertains to Bayesian inference. We show,
experimentally, how misinterpretations of uncertain evidence can lead to vastly
different inference results; emphasizing the importance of carefully accounting
for uncertain evidence.

\section{Background}\label{sec:background}

Bayesian inference aims to characterize the posterior distribution of the latent
random vector $\bix$ given the observed random vector $\biy$. When observing
$\biy$ with certainty the inference problem is ``straightforward'' in the sense
that $p(\bix|\biy)=p(\biy,\bix)/p(\biy)$. However, exact inference is often
infeasible as $p(\biy)$ is usually intractable, but if the joint $p(\biy,\bix)$
is calculable then inference is achievable via approximate methods such as
importance sampling~\citep[e.g.][]{hammersley1964monte}, Metropolis-Hastings
\citep{metropolis1949monte,metropolis1953equation,hastings1970monte}, and
Hamiltonian Monte Carlo~\citep{duane1987hybrid,NEAL1994194}. Unfortunately,
standard Bayesian inference is incompatible with uncertain evidence where exact
values of $\biy$ are unavailable.

Before discussing ways to treat uncertain evidence, we first introduce the
highest level abstraction representing uncertain evidence. Specifically, we
consider $\epsilon\in\mathcal{E}$, where $\mathcal{E}$ is a set of
``statements'' specifying the uncertainty about $\biy$. For example, in the drop
of a ball example $\epsilon$ would be a statement represented as
\cref{tab:times-uncertain}. In contrast $\zeta$ is a lower level abstraction
which is encoded in $\epsilon$. Dealing with uncertain evidence is a matter of
decoding or interpreting $\epsilon$, possibly identifying $\zeta$ and relating
it to $p(\biy,\bix)$. The canonical example of interpreting uncertain evidence,
as introduced by \citep[p. 165]{jeffrey1965logic}, is ``observation by
candlelight,'' which motivated \textit{Jeffrey's rule}:

\bdef[Jeffrey's Rule \citep{jeffrey1965logic}]\label{def:jeffreys-rule} Given
$p(\biy,\bix)$, let the interpretation of a given $\epsilon \in\mathcal{E}$ lead
to $\biy$ being associated with uncertainty, conditioned on auxiliary evidence
$\zeta$---where $\zeta$ may be unknown---and denote the decoded uncertainty by
$q(\biy|\zeta)$. Then the updated (posterior) distribution $p(\bix|\zeta)$ is:
\begin{equation}
  \label{eq:jeffrey}
  p(\bix|\zeta) = \EE_{q(\biy|\zeta)} \left[ p(\bix | \biy) \right].
\end{equation}
In particular, one considers the updated joint $p(\biy,\bix|\zeta)=p(\bix | \biy)
q(\biy|\zeta)$, such that $q(\biy|\zeta)$ is a marginal of $p(\biy,\bix | \zeta)$.
\edef
Jeffrey envisioned the existence of the auxiliary variable (or vector), $\zeta$;
however, Jeffrey's rule is often defined without it
\citep[e.g.,][]{chan2005revision}. Nonetheless, we argue that reasoning about an
auxiliary variable (or vector) $\zeta$ is the more intuitive perspective as
\textit{some} evidence must have given rise to $q$. Further, accompanying the
introduction of Jeffrey's rule is the preservation of the conditional
distribution of $\bix$ upon applying Jeffrey's rule, see
e.g.~\cite{jeffrey1965logic,diaconis1982updating,valtorta2002soft}
and~\cite{chan2005revision}. That is, the evidence $\zeta$ giving rise to
$q(\biy|\zeta)$ must not also alter the conditional distribution of $\bix$ given
$\biy$. Mathematically, Jeffrey's rule requires that,
$p(\bix|\biy,\zeta)=p(\bix|\biy)$. This, for instance, relates to the
commutativity of Jeffrey's rule, which is treated in full detail
by~\citet{diaconis1982updating}, and briefly discuss in \cref{app:comm-jeff}.

In contrast to Jeffrey's rule is \textit{virtual evidence} as proposed
by~\citet{pearl1988probabilistic}. Virtual evidence also includes an auxiliary
\textit{virtual} variable (or vector), but does so via the likelihood
$q(\zeta|\biy,\bix)\defas q(\zeta|\biy)$, with the only parents of $\zeta$ being
$\biy$:

\bdef[Virtual evidence \citep{pearl1988probabilistic}]\label{def:virtual-evidence} 

Given $p(\biy,\bix)$ and suppose a given $\epsilon \in\mathcal{E}$ leads to the
interpretation that we extend $p(\biy,\bix)$ with an auxiliary virtual variable
(or vector) $\zeta$ such that: (1) in the discrete case, where the values of
$\biy\in\left\{ \biy_{k} \right\}_{k=1}^{K}$ are mutually exclusive, the
uncertain evidence is decoded to as likelihood ratios\footnote{The notation for
ratios containing several terms, for example A, B, and C, is written as $x:y:z$.
This is understood as: ``for every $x$ part of A there is $y$ part B and $z$
part C.''} $\left\{ \lambda_{k} \right\}_{k=1}^{K}$:
\begin{equation}
  \label{eq:lik-ratios}
  \lambda_{1} : \cdots : \lambda_{K} = q(\zeta|\biy_{1}) : \cdots :
  q(\zeta|\biy_{K}).
\end{equation}
The posterior over $\bix$ given uncertain evidence is
(\citealt{chan2005revision}; a result we also prove in
\cref{app:proof-virtual}),
\begin{equation}
  \label{eq:pearls-discrete}
  p(\bix|\zeta) = \frac{\sum_{k=1}^K \lambda^{k}p(\biy_{k},\bix)}{\sum_{j=1}^K \lambda^{j}p(\biy_{j})}\ .
\end{equation}
(2) If $\biy$ is continuous, decoding $\epsilon$ leads to the virtual likelihood
$q(\zeta|\biy)$ such that the posterior is proportional to the (virtual) joint
\[ \label{eq:pearls-continuous}
  p(\bix|\zeta) \propto \int p(\zeta,\biy,\bix)\dd \biy =\int q(\zeta|\biy)p(\biy,\bix)\dd \biy.
\]
\edef
\begin{figure}[t!]
  \centering
  \includegraphics{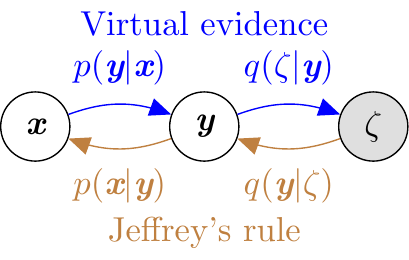}
  \caption{\label{fig:jeffrey-vs-pearl} Jeffrey's rule compared to virtual
    evidence in terms of the auxiliary evidence $\zeta$. Both virtual evidence
    and Jeffrey's rule are defined in terms of the base model $p(\biy,\bix)$.}
\end{figure}
In practice, in the continuous case one can approximate the posterior using
standard approximate inference algorithms requiring only the evaluation of the
joint. In the discrete case, Eq.~\ref{eq:pearls-discrete}, the posterior
inference is exact assuming a known $p(\biy_{i})$ for all $i\in
\left\{1,\dots,K\right\}$. When comparing Jeffrey's rule and virtual
evidence~\citep[e.g.,][]{pearl1988probabilistic,valtorta2002soft,jacobs2019mathematics}
we can do so in terms of $\zeta$ and the corresponding graphical model
(\cref{fig:jeffrey-vs-pearl}). This figure is a graphical representation of how
Jeffrey's rule and virtual evidence relate $\zeta$ to the existing probabilistic
model, $p(\biy,\bix)$. Particularly, Jeffrey's rule and virtual evidence affect
the model in \textit{opposite} directions. Jeffrey's rule pertains to
uncertainty about $\biy$ \textit{given} some evidence, while virtual evidence
requires reasoning about the likelihood $q(\zeta|\biy)$.

It is (perhaps) not surprising that one may apply Jeffrey's rule, yet implement
it as a special case of virtual evidence, by choosing a particular form of
likelihood ratios, \cref{eq:lik-ratios}, and vice versa
\citep{pearl1988probabilistic,chan2005revision}. However, this is of purely
algorithmic significance as the two approaches remain fundamentally different.

A third approach to uncertain evidence, recently introduced
by~\cite{tolpin2021probabilistic}, treats the uncertain evidence on $\biy$ as an
event. This approach, which we refer to as \textit{distributional evidence},
defines a likelihood on the event $\left\{ \biy \sim D_q \right\}$ (reads as
``the event that the distribution of $\biy$ is $D_q$ with density $q(\biy)$'')
and considers the auxiliary variable $\zeta=\left\{ \biy \sim D_q \right\}$:
\bdef[Distributional evidence
\citep{tolpin2021probabilistic}]\label{def:distributional-evidence} 

Let $p(\biy,\bix)=p(\biy|\bix)p(\bix)$ be the joint distribution with a known
factorization. Assume the interpretation of a given $\epsilon\in \mathcal{E}$
yields a density $q(\biy)$, with distribution $D_q$. Define the likelihood
$p(\biy\sim D_q | \bix)$ as:
\begin{equation}
  \label{eq:distributional-evidence}
  p(\zeta|\bix) =
  \frac{\exp\EE_{q(\biy)}\left[ \ln  p(\biy|\bix) \right]}{Z(\bix)}
\end{equation}
where $\zeta= \left\{\biy\sim D_q\right\} $ and $Z(\bix)$ is a normalization
constant that generally depends on $\bix$. Typically we drop explicitly writing
$\zeta$ and simply write $q(\biy \sim D_q|\bix)$. See
\citep{tolpin2021probabilistic} for sufficient conditions for which
$Z(\bix)<\infty$.
\edef

\section{Which Approach?}

The lack of a general consensus on how best to approach uncertain evidence means
that it is difficult to know what to do, in practical terms, when faced with
uncertain evidence. In isolation, each approach discussed in the previous
section appears well supported, even when applied to the same model
\citep[e.g.,][]{benmrad2013understanding}. However, the underlying arguments
remain somewhat circumstantial. Prior work tends to create contexts tailored for
each approach and it is unclear how relatable or generalizable those contexts
are. As such, much prior work is not particularly instructive when deducing
which approach to adopt for new applications that do not fit those prior
context. We argue that the apparent philosophical discourse fundamentally stem
from a disagreement about the model $M\in \tilde{\mathcal{M}}$ in which we seek
to do inference given uncertain evidence $\epsilon\in\mathcal{E}$. This can be
framed as an inference problem where we seek to find (or directly define)
$p(M|\epsilon)$. The significance of this perspective is that reasoning about
the triplet $M\in \mathcal{M}$, $\epsilon\in\mathcal{E}$, and $p(M|\epsilon)$
makes for a better foundation that encourages discussions about and makes clear
the underlying assumptions.

How then should we define $p(M|\epsilon)$? In the general case, reaching
consensus is close to impossible as it requires fully specifying $\mathcal{M}$
and $\mathcal{E}$ (all possible models and conceivable evidences). However,
while universal consensus is arguably unattainable; ``local'' consensus might
be. Here locality refers to defining $p(M|\epsilon)$ on constrained and
application dependent subsets $\tilde{\mathcal{E}} \subset \mathcal{E}$ and
$\tilde{\mathcal{M}} \subset \mathcal{M}$. This perspective was considered
by~\cite{grove1997probability}, yet does not seem to have resurfaced in this
context since.~\citet{grove1997probability} define $\tilde{\mathcal{M}}$ in
terms of a prior $p(M)$ and implicitly defines $\tilde{\mathcal{E}}$ as a set of
trusted statements pertaining to (conditional) probabilities. They further
define the likelihood $p(\epsilon|M)$ which evaluates to one if the model $M$ is
consistent with the evidence $\epsilon$ and zero otherwise. From this they are
able to calculate $p(M|\epsilon)\propto p(\epsilon|M)p(M)$. 

\subsection{Uncertain Evidence Interpretation}\label{sec:uncertain-evid-interp}

We propose to limit the consideration of $\tilde{\mathcal{E}}$ and
$\tilde{\mathcal{M}}$ to constrained, but widely applicable (and application
dependent) subsets set in the context of inference. To construct
$\tilde{\mathcal{E}}$ and $\tilde{\mathcal{M}}$ we begin with the assumption
that a \textit{base model}, $p(\biy,\bix)$, is always available. We further
assume that $\tilde{\mathcal{E}}$ contains evidence in the form of statements
which we interpret in a literal sense. To ensure inference with exact evidence
is possible, we require that $\tilde{\mathcal{E}}$ contain evidences that encode
exact evidence about $\biy$. For example $\epsilon=\,$~``the value of $\biy$ is
$\hat{\biy}$.'' Finally, we constrain the form of \textit{uncertain evidence} by
requiring $\epsilon$ to encode uncertainty in one of three ways: (I) $\epsilon$
encodes a distribution $q$ over $\biy$, for example $\epsilon=\,$~``The
distribution of $\biy$ is $q(\biy|\zeta)$''. (II) $\epsilon$ encodes a
\textit{conditional} distribution about $\biy$ given $\bix=\hat{\bix}$, for
example $\epsilon=\text{``iff}~\bix=\hat{\bix}~\text{then the distribution
of}~\biy~\text{is}~q(\biy|\bix=\hat{\bix}).\text{''}$ (III) Uncertain evidence
is explicitly expressed in terms of a likelihood of $\biy$, for example let
$\biy\in \left\{ 0,1 \right\}$ and consider $\epsilon=\,$~``$\biy=1$ is twice as
likely to explain the evidence compared to $\biy=0$.'' We define
$\tilde{\mathcal{M}}$ implicitly by requiring that the random variable
$\epsilon$ partitions $\tilde{\mathcal{M}}$ such that the posterior
$p(\bix|\epsilon)$ takes a certain form:
\bdef Given $\epsilon\in\tilde{\mathcal{E}}$, we define
$p(\tilde{\mathcal{M}}|\epsilon)$ and $\tilde{\mathcal{M}}$ implicitly through
the partitions of $\tilde{\mathcal{M}}$ as generated by $\epsilon$, such that
inference given $\epsilon$ becomes,
\begin{align}
   p(\bix|\epsilon) &=\EE_{p(M|\epsilon)}[p(\bix|M)] \\
   & = \begin{cases}
         p(\bix|\biy), & \text{if}~\epsilon~\text{is exact},\\
         \int p(\bix|\biy)q(\biy|\zeta) \dd\biy, & \text{if}~\epsilon~\text{is type (I) }, \\[0.4em]
         \frac{p(\bix)q(\biy\sim D_q|\bix)}{p(\biy\sim q)}, & \text{if}~\epsilon~\text{is type (II) }, \\[0.4em]
         \frac{\int p(\bix)p(\biy|\bix)q(\zeta|\biy)\dd\biy}{p(\zeta)}, & \text{if}~\epsilon~\text{is type (III) },
       \end{cases}
  \label{eq:model-expectation}
\end{align}
\edef
where type (I-III) leads to Jeffrey's rule, distributional evidence, and virtual
evidence respectively. We emphasize, that the definitions of
$\tilde{\mathcal{E}}$, $\tilde{\mathcal{M}}$, and
$p(\tilde{\mathcal{M}}|\epsilon)$ are \textit{not} fundamental truths. Rather,
they are subject to our beliefs about how one ought to approach uncertain
evidence in a form found in $\tilde{\mathcal{E}}$. In particular, notice that
type (I) and (II) evidences are similar in that they describe a distribution of
$\biy$. The crucial difference lies in the conditional relationship giving rise
to said probability. In type (I) we assume uncertainty is due to external
(unknown) evidence, represented by $\zeta$ not found in $\bix$ or $\biy$. On the
other hand, in type (II) $\zeta$ \textit{is} $\bix$ (or a subset thereof). Even
though we argue that Jeffrey's rule is preferable given type (I) uncertain
evidence, it turns out there are cases where Jeffrey's rule is, in fact,
inconsistent with $p(\biy,\bix)$---which we show in the following section.
Nonetheless, from a mathematical perspective, Jeffrey's rule can still be
applied. This is justified, in part, as Jeffrey's rule leads to a ``new'' model
$p(\biy,\bix|\zeta)$ which is closest to $p(\biy,\bix)$ as measured by the KL
divergence $\kl{p(\biy,\bix|\zeta)}{p(\biy,\bix)}$ constrained such that $\int
p(\biy,\bix|\zeta) \mathrm{d}\bix = q(\biy|\zeta)$ \citep[and citations
therein]{peng2010bayesian}. Despite this, if Jeffrey's rule is inconsistent with
$p(\biy,\bix)$ it may preferable to either: (1) update the model $p(\biy,\bix)$
to be compatible with the given uncertain evidence or (2) acquire compatible
data---be it exact observations or better uncertain evidence.

\subsection{Consistency}\label{sec:consistency}

We define consistency in terms of whether or not one can extend the joint
distribution with auxiliary variables (or vectors) such as to contain the
uncertainty encoded in $\epsilon\in \tilde{\mathcal{E}}$,

\bdef[Consistency]

Consider an auxiliary variable (or vector) $\zeta$ and the associated density
$q$ derived from $\epsilon$, where $q$ can take the form of either
$q(\zeta|\cdot)$ or $q(\cdot|\zeta)$. We then say that Jeffrey's rule, virtual
evidence, and distributional evidence are consistent with $p(\biy,\bix)$ if a
joint exists, $p(\zeta,\biy,\bix)=p(\zeta|\biy,\bix)p(\biy,\bix)$, such that
either $p(\zeta|\cdot)=q(\zeta|\cdot)$ or $p(\cdot|\zeta)=q(\cdot|\zeta)$
depending on the form of $q$.
\edef
Both virtual evidence and distributional evidence are, by their definition,
always consistent. Virtual evidence is defined as an extension of the graphical
model $p(\biy,\bix)$ through the auxiliary variable (or vector) $\zeta$ and its
likelihood $q(\zeta|\biy)$. That is we can always consider $p(\zeta|\biy)\defas
q(\zeta|\biy)$ such that $p(\zeta,\biy,\bix)\defas p(\zeta|\biy)p(\biy,\bix)$.
Similarly, in the case of distributional evidence, we can consider
$p(\zeta,\biy,\bix) \defas p(\zeta|\bix)p(\biy|\bix)p(\bix)$. However, despite
distributional evidence being consistent, notice that it introduces $\zeta$ as
independent of $\biy$. As such, distributional evidence introduces an entirely
new likelihood with respect to $\bix$ and we can consider $p(\zeta|\bix)p(\bix)$
as a \textit{new} model. This results in the loss of the physical interpretation
of the relationship between $\biy$ and $\bix$ as defined through $p(\biy|\bix)$
even though $q(\zeta|\bix)$ is derived from $p(\biy|\bix)$. On the other hand,
in the case of Jeffrey's rule, we cannot guarantee consistency, and so one needs
to be mindful of the potential mismatch between the base model and
$q(\biy|\zeta)$. While~\cite{diaconis1982updating} provide an extensive and
theoretical examination of Jeffrey's rule they leave out important points
concerning necessary conditions for Jeffrey's rule to satisfy consistency that
we present here and prove in \cref{app:proofs-consistent}:
\bthm \label{thm:consistency-jeffrey} Necessary and sufficient conditions for
Jeffrey's rule to be consistent with respect to $p(\biy,\bix)$ given
$q(\biy|\zeta)$ (that is, there exists a joint $p(\zeta,\biy,\bix)$ such that
$p(\biy|\zeta) = q(\biy|\zeta)$):

\begin{enumerate}
  \item (Necessary and sufficient) There exists $p(\zeta|\biy)$ such that for
        all $\zeta$ and $\biy$,
\[
          q(\biy|\zeta) = \frac{p(\zeta|\biy)p(\biy)}{\EE_{p(\biy)} \left[
              p(\zeta|\biy) \right]}
        \]
  \item (Necessary) If $q(\biy|\zeta)=\prod_{i=1}^{D}q(y_{i}|\zeta)$ then it
        must hold that: (1) $\zeta$ is a random vector $\zeta=\left(
        \zeta_{1},\dots,\zeta_{D}\right)$ where each $\zeta_{i}$ uniquely links
        to $y_{i}$ such that $q(y_{i}|\zeta)=q(y_{i}|\zeta_{i})$ and (2) $\bix$
        is likewise multivariate and each $x_i$ uniquely links to $y_{i}$ such
        that $p(y_{i}|\bix)=p(y_{i}|x_{i})$.
  \item (Necessary) Let $p(\zeta)=\EE\left[ p(\zeta|\biy) \right]$, then it must
        hold that: (1) $\cov\left[ \biy \right] \succeq \EE\left[ \cov\left[
        \biy |\zeta \right] \right] $, where $\succeq$ denotes determinant
        inequality. (2) For each $y_{i}$ it holds that $\var\left[ y_{i} \right]
        \geq \EE\left[ \var\left[ y_{i} | \zeta \right] \right] $. In
        particular, if the variance $\var\left[ y_{i} |\zeta \right]=\sigma^{2}$
        is constant and independent of $\zeta$ we have $\var\left[ y_{i} \right]
        \geq \sigma^{2} $ with equality if and only if $\EE\left[ y_i | \zeta
        \right] = \mu$ is constant.
\end{enumerate}
\ethm
Unfortunately, validating consistency of Jeffrey's rule is in general infeasible
as \cref{thm:consistency-jeffrey} (1) is usually intractable to assess. One can
only reliably conclude if Jeffrey's rule is inconsistent in special cases via
\cref{thm:consistency-jeffrey} (2-3). 

\subsection{Distributional Evidence: Exact or Implied Inference?}

While we generally prefer Jeffrey's rule over distributional evidence and
although Jeffrey's rule is technically applicable given type (II) uncertain
evidence, why then do we prefer distributional evidence given type (II)? If we
were to use Jeffrey's rule in this case its interpretation becomes unclear if
$q$ is of the form $q(\biy | g(\bix))$ where $g(\cdot)$ is a selector function
which selects a subset of the variables in $\bix$. As we ultimately seek to
infer a posterior over the latent variables given $\zeta=g(\bix)$, it violates
the intuition that $\zeta$ should be an auxiliary variable (or vector) not found
in $\bix$, which is required by Jeffrey's rule. Specifically, we can consider
two kinds of uncertain evidence of this form: (1) a functional $q(\biy |
g(\bix))$ specified for all $\bix$ and $\biy$ and (2) a conditional form such
that $q$ is a distribution specified for only a specific value of $g(\bix) =
g(\hat{\bix})$. In case (1) one arguably ought to replace the model
$p(\biy,\bix) \rightarrow q(\biy|g(\bix))p(\bix)$ such that $q(\biy|g(\bix))$
becomes the new likelihood. However, in case (2) we cannot simply replace the
model, as we do not know the form of $q$ for any other value of $g(\bix)$ than
$g(\hat{\bix})$. In particular, we can think of case (2) as the limiting case of
observing $\mathcal{D}=\left\{ \biy_{i} \right\}_{i=1}^{N}$ for
$N\rightarrow\infty$, where $\biy_{i}\overset{i.i.d.}{\sim} p(\biy
|g(\bix)=g(\hat{\bix}))$, where the empirical distribution of $\mathcal{D}$ in
the limit represents $p(\biy|g(\bix)=g(\hat{\bix}))$. As pointed out also
by~\citet{tolpin2021probabilistic} there is a similarity between observing
$\mathcal{D}$ for large $N$ and instead condition on $q(\biy)$ associated with
the empirical distribution represented by $\mathcal{D}$. From this perspective,
distributional evidence provides for inferring $p(\bix | \biy \sim D_q )$ as
opposed to $p(\bix | \mathcal{D})$. This view is useful, particularly when
$\mathcal{D}$ is unavailable yet its distributive representation, $q$, is. One
caveat to distributional evidence, that~\citet{tolpin2021probabilistic} do not
discuss, is whether or not $Z(\bix)$ in \cref{eq:distributional-evidence} is
calculable. In particular,~\citet{tolpin2021probabilistic} appears to leave it
as a normalization constant that is never calculated. That is, they compute the
function $f(\biy \sim D_q | \bix)=p(\biy \sim D_q | \bix)Z(\bix)$ when
performing inference, where $f$ is the numerator in
\cref{eq:distributional-evidence}---a ``pseudo-likelihood.'' The difference
between computing $p(\biy \sim D_q | \bix)$ and $f(\biy \sim D_q | \bix)$ in the
context of inference is:
\[
  p(\bix | \biy \sim D_q) \propto \begin{cases}
                      p(\biy \sim D_q | \bix)p(\bix) & \text{if known}~Z(\bix), \\
                      f(\biy \sim D_q | \bix)p(\bix) & \text{otherwise}.
             \end{cases}
\]
While the first expression above leads to posterior inference as expected, the
second expression leads to an implied posterior via the implied joint:
\begin{align}
  f(\biy \sim D_q | \bix)p(\bix) &= p(\biy \sim D_q | \bix)p(\bix)Z(\bix) \\
                                 & = p(\biy \sim D_q | \bix)\hat{p}_{\mathrm{a}}(\bix),
  \label{eq:distributional-evidence-implied-joint}
\end{align}
where $\hat{p}_{\mathrm{a}}(\bix)=p(\bix)Z(\bix)$ is a \textit{distributional
evidence adjusted} unormalized prior on $\bix$. As such, regardless of whether
or not a known $Z(\bix)$ is available, the same likelihood on the event $
\left\{ \biy \sim D_q \right\}$ but different priors on $\bix$ is used. To
ensure that the use of \cref{eq:distributional-evidence-implied-joint} leads to
a valid posterior, it is enough to show that the adjusted prior
$p_{\mathrm{a}}(\bix)\propto \hat{p}_{\mathrm{a}}(\bix)$ normalizes in $\bix$:

\bthm \label{thm:stoch-evid-adjusted-prior} Under the same assumptions as in
Theorem 1 in the paper of~\cite{tolpin2021probabilistic}, the adjusted prior
$p_{\mathrm{a}}(\bix) = p(\bix)Z(\bix)/C$ normalizes. That is $C
< \infty$.
\ethm
\bprfof{~\cref{thm:stoch-evid-adjusted-prior}} Assume, as done
by~\cite{tolpin2021probabilistic}, that the set of distributions $\mathcal{Q}$
is implicitly defined through the set of parameters $\Theta$ where
$\theta\in\Theta$ parameterizes $q_{\theta}$ such that $\mathcal{Q}= \left\{
q_{\theta} | \theta \in \Theta\right\}$. Assume further that $\sup_{\biy}
\int_{\Theta}q_{\theta}(\biy)\dd \theta < \infty$.  Then the bound on
$Z(\bix)$, as derived by~\cite{tolpin2021probabilistic}, is independent of
$\bix$. It then follows that $Z(\bix) \leq \tilde{Z}$ for all $\bix$ such
that:
\begin{align}
  C & = \int \hat{p}_{\mathrm{a}}(\bix) \dd\bix =  \int p(\bix) Z(\bix) \dd\bix \\ 
    & \leq \int p(\bix) \tilde{Z} \dd\bix = \tilde{Z} < \infty.
\end{align}
This implies that $p_{\mathrm{a}}(\bix)=p(\bix)Z(\bix)/C$ is a valid marginal
as it normalizes in $\bix$, which concludes the proof.
\eprfof
\subsection{Complexity}

The primary consideration when comparing Jeffrey's rule, virtual evidence, and
distributional evidence, is their applicability given a certain type of
uncertain evidence. In a practical setting, it is unclear by how much each
approach differs in their posteriors over $\bix$ given the same uncertain
evidence. As we illustrate in \cref{sec:uncert-evid-mult}, this difference may
range from significant to negligible and is a function of the base model as well
as the uncertain evidence. Therefore, when inference is time-sensitive, it may
be beneficial to initially perform inference using an approach of low
computational complexity and then subsequently follow up with the appropriate
approach to verify inference results. Our complexity analysis assumes no
analytical solution is feasible, and that approximate inference is employed;
that is, sampling-based inference methods as well as Monte Carlo estimations of
expectations is used. We use $c_{\mathrm{i}}$ to denote the complexity for
achieving adequate approximate posterior inference, and we use $n_{\mathrm{e}}$
to denote the number of required samples for adequate Monte Carlo estimations of
expectations. Note that this relies on the additional assumption that inferring
$p(\bix|\biy)$ has the same complexity as inferring $p(\bix|\zeta)=\int
p(\bix,\biy|\zeta)\,\mathrm{d}\biy$. From this we find that the complexity of
Jeffrey's rule, \cref{eq:jeffrey}, requires estimating the expected posterior
leading to a complexity of $c_{\mathrm{i}}n_{\mathrm{e}}$, while virtual
evidence is $c_{\mathrm{i}}$ as it only involves inferring the posterior under
the joint given by \cref{eq:pearls-continuous}. As for distributional evidence,
\cref{eq:distributional-evidence}, if the new likelihood is analytically
tractable the complexity is $c_{\mathrm{i}}$, since it requires only inferring a
posterior distribution. If the likelihood is approximated using Monte Carlo
estimation the complexity increases to $c_{\mathrm{i}}n_{\mathrm{e}}$.
Therefore, virtual evidence is, in general, more efficient than both Jeffrey's
rule and distributional evidence.

Finally, we note that a reduction of the complexity gap between Jeffrey's rule
and virtual evidence is achievable using amortized
inference~\cite{gershman2014amortized}. Amortized inference reduces the cost of
inference in exchange for an upfront computational cost. Therefore, estimating
an expected posterior, which is the case for Jeffrey's rule, can be
significantly sped up.

\begin{figure}[t!]
  \centering
  \includegraphics[width=\columnwidth]{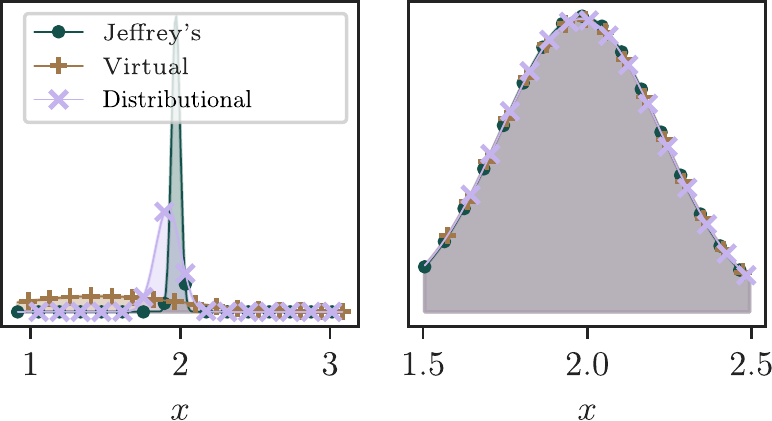}
  \caption{Analytical posterior results given uncertain evidence $q(y|\zeta)$
    using Jeffrey's rule, virtual evidence, and distributional evidence. (Left) We
    set $\mu_x=1$, $\sigma_x=1$, $\sigma_{y|x}=0.3$, $\sigma_q=1$, and $\zeta=2$
    from which we derive the remaining means and (conditional) variances as
    described in \cref{sec:uncert-evid-mult}. (Right) same as (left) except with
    $\mu_x=0$, $\sigma_x=5$, $\sigma_{y|x}=0.5$, $\sigma_q=0.5$, and $\zeta=2$.
    }
  \label{fig:multi-var-results}
\end{figure}

\section{Experiments}\label{sec:experiments}

In this section we illustrate the importance of making the appropriate
interpretation and treatment of uncertain evidence. We carry out three
experiments in which the appropriate treatment of the given uncertain evidence
is to use Jeffrey's rule. We then compare against making a
\textit{misinterpretation} leading to either virtual evidence or distributional
evidence. We demonstrate how such misinterpretations can lead to inference
results that range from being significantly different to almost
indistinguishable. Most prior
work~\citep[e.g.,][]{chan2005revision,benmrad2013understanding,mrad2015explication,jacobs2019mathematics}
compares only Jeffrey's rule and virtual evidence for discrete problems, whereas
we focus on the continuous case.

\subsection{Uncertain Evidence and the Multivariate
  Gaussian}\label{sec:uncert-evid-mult}

We consider a multivariate Gaussian model where the base model factorizes as
$p(x,y)=p(x)p(y|x)$ where $p(x)=\distNorm(\mu_{x},\sigma_{x}^{2})$ and
$p(y|x)=\distNorm(x,\sigma_{y|x}^2)$. The aim is to infer the posterior
distribution of $x$, ideally given an exact observation of $y$. However, we
assume this is unavailable and instead we are given uncertain evidence,
$\epsilon$, of type (I); that is we are given the density
$q(y|\zeta)=\distNorm(\zeta,\sigma_{q}^2)$ and instead seek to infer
$p(x|\zeta)$. Using \cref{eq:model-expectation} implies performing inference
using Jeffrey's rule. To ensure Jeffrey's rule is consistent,
\cref{thm:consistency-jeffrey}, we take on the perspective of an ``oracle'' and
impose the restriction that all marginal and conditionals are Gaussians, which
leads to the joint also being Gaussian~\citep[e.g.][ch. 2.3]{bishop2006pattern}.
We see that \cref{thm:consistency-jeffrey} (II) is trivially satisfied as
$\bix=x$, $\biy=y$, and $\zeta$ are one-dimensional. Further, we find a
$p(\zeta|y)$ that satisfies \cref{thm:consistency-jeffrey} (I) by choosing
$p(\zeta|y)=\distNorm(\mu_{\zeta|y},\sigma_{\zeta|y}^2)$ such that
$\mu_{\zeta|y}=(y\sigma_\zeta^2 +
\mu_x\sigma_{q}^2)/(\sigma_{\zeta}^2+\sigma_{q}^2)$ and
$\sigma_{\zeta|y}^2=\sigma_\zeta^2\sigma_{q}^2/(\sigma_{\zeta}^2+\sigma_{q}^2)$
where $\sigma_{\zeta}^2 = \sigma_{x}^2 + \sigma_{y|x}^2 - \sigma_{q}^2$.
Specifically, we see how the variance constraint $\sigma_\zeta^2\geq0$ ensures
that we satisfy \cref{thm:consistency-jeffrey} (III) as
$\sigma_{\zeta}^2\geq0\Rightarrow\sigma_y^2=\sigma_{x}^2 +
\sigma_{y|x}^2\geq\sigma_q^2=\EE\left[\var\left[y|\zeta\right]\right]$. Please
see~\cref{fig:multi-var-results} for the values used in the experiment.

\begin{figure}[t!]
  \centering
  \includegraphics{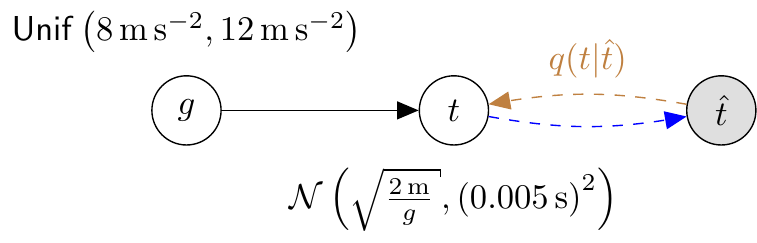}
  \caption{\label{fig:drop-ball-graph} Graphical model of the \textit{drop of a
    ball} experiment in \cref{sec:ball-drop}. We use a brown dashed edge to
    specify the posited uncertain evidence density, while the blue dashed edge
    emphasizes that we do not know the true $p(\hat{t}|t)$. Rather, the blue
    edge represent \textit{interpreting} uncertain evidence of type (I) as type
    (III) leading to virtual evidence.}
\end{figure}

When comparing Jeffrey's rule to virtual and distributional evidence we fix the
base model $p(x,y)$ and the density $q(y|\zeta)$ but vary the interpretation of
the distributional evidence. In particular, since $q(y|\zeta)$ is symmetric in
$y$ and $\zeta$, we take for virtual evidence $q_{\mathrm{V}}(\zeta|y) =
\distNorm(y,\sigma_{q_\zeta}^2)$. In the case of distributional evidence we
analytically solve for $p(y\sim D_q|x)$ as well as the adjusted prior
$p_{\mathrm{a}}(x)$ by assuming the density $D$ is implicitly defined through
the mean, $\theta$, of $q(y|\zeta)=\distNorm(\theta,\sigma_q^2)$, see
\cref{app:distributional-evi-form}. In all cases, we arrive at Gaussian
distributed posteriors $p(x|\zeta)$, and we show the different posteriors in
\cref{fig:multi-var-results}. We note how in \cref{fig:multi-var-results}, in
the left panel the three methods result in vastly different posteriors, whereas
those in the right panel are indistinguishable. This emphasizes the importance
of carefully choosing the approach in dealing with uncertain evidence.

\subsection{The Drop of a Ball}\label{sec:ball-drop}

Consider the running example in this paper of the classic ``high school''
experiment in which a student attempts to measure gravitational acceleration,
$g$, by timing, $t$, how long it takes for a ball to fall a distance, $x$. Armed
with the formula $x=g t^2/2$ our student can convert measurements of $t$ into
estimates of $g$ if $x$ is known. In our setup $x=\SI{1}{\meter}$ and is
measured a-priori, we assume a ``model'' error of $\SI{0.005}{\second}$ to
account for physics ignored by our formula (e.g., air resistance) and we assume
an error of $\SI{0.03}{\second}$ on the true time, $t$, given the observation,
$\hat{t}$, produced by the stopwatch (type (I) uncertain evidence)---our student
does not trust their ability to hit the ``stop'' button as the ball hits the
ground more accurately than this. Our (lazy) student then attempts to infer $g$
from a single experiment, during which they observe a time on the
stopwatch\footnote{A perfect experiment would record $\simeq 0.45\,\mathrm{s}$
for the terrestrial $g\simeq\SI{9.81}{\meter\per\second\squared}$.} of
$\SI{0.43}{\second}$, leading to
$q(t|\hat{t})=\distNorm(\SI{0.43}{\second},\left(\SI{0.03}{\second}\right)^2)$.
We show the graphical model in \cref{fig:drop-ball-graph}. 

\begin{figure}[t!]
  \centering
  \includegraphics[width=\columnwidth]{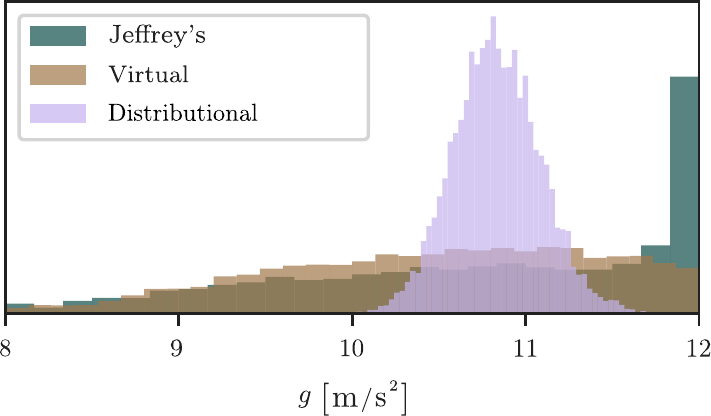}
  \caption{Posterior distributions over the gravitational acceleration, $g$, on
  the surface of Earth inferred by an experiment in which the time taken,
  $\hat{t}=\SI{0.43}{\second}$, for a ball to fall $\SI{1}{\meter}$ is measured.
  Given the uncertain evidence
  $q(t|\hat{t})=\distNorm(\SI{0.43}{\second},\left(\SI{0.03}{\second}\right)^2)$
  we notice that $g\simeq\SI{9.81}{\meter\per\second\squared}$ is well covered
  by the posteriors of Jeffrey's rule and virtual evidence but is excluded by
  distributional evidence. Recall, however, that both virtual evidence and
  distributional evidence are \textit{inappropriate} in this case and serve to
  illustrate what happens when the uncertain evidence is misinterpreted.}
  \label{fig:ball}
\end{figure}

For virtual evidence we again `flip' $q$, as it is symmetric in its mean and
random variable, such that
$p_\mathrm{V}(\hat{t}|t)=\distNorm(t,\left(\SI{0.03}{\second}\right)^2)$. For
distributional evidence we notice the form of $q(t\sim D_q|g)$ is the same as in
\cref{sec:uncert-evid-mult} which allows for an analytical likelihood. In the
case of Jeffrey's rule we similarly trivially satisfy
\cref{thm:consistency-jeffrey} (II), as $g$, $t$, and $\hat{t}$ are
one-dimensional, as well as \cref{thm:consistency-jeffrey} (III), as $\var[y]
=\EE\left[\var[t|g]\right] +
\var[\EE[t|g]]=0.005^2+\EE[2/g]-(\EE[\sqrt{2}/\sqrt{g}])^2=0.0007\geq 0.003^2 =
\var[\EE[t|\hat{t}]]$. We also note that different to the experiment in
\cref{sec:uncert-evid-mult} we do not compute analytical posteriors but infer
those using approximate Bayesian inference via the probabilistic programming
language \textsc{pyprob}~\citep{pyprob}. \cref{fig:ball} shows the posteriors
using the three different interpretations of the given uncertain evidence. We
see that each posterior is different with Jeffrey's rule and virtual evidence
being more similar compared to distributional evidence. In particular we note
the small variance of distributional evidence results in near zero probability
on the true value of the gravitational acceleration at
$g\simeq\SI{9.81}{\meter\per\second\squared}$. 

This again exemplifies the potential error one might make when a certain type of
uncertain evidence is misinterpreted. Particularly, distributional evidence
should not be expected to produce reasonable results in this case. Recall that
both virtual and distributional evidence are inappropriate by construction. To
make, for example, distributional evidence the correct interpretation, we can
modify this example so that the student's setup is somewhat shaky, so $x$ varies
slightly. The student instead uses a very accurate time measurement device; the
measurements of the time are \textit{exact}. The student may then carry out
repeated measurements in this \textit{single} experiment and conclude that the
measured time $t$ is distributed as $q(t|g)=\distNorm(\mu_t, \sigma_t^2)$. In
this case the uncertain evidence is of type (II) since the uncertainty is
conditioned on the latent variable $g$.

\subsection{Planet Orbiting Kepler 90}
\label{sed:orbiting-kepler}

\begin{figure}[t!]
  \centering
  \includegraphics[width=\columnwidth]{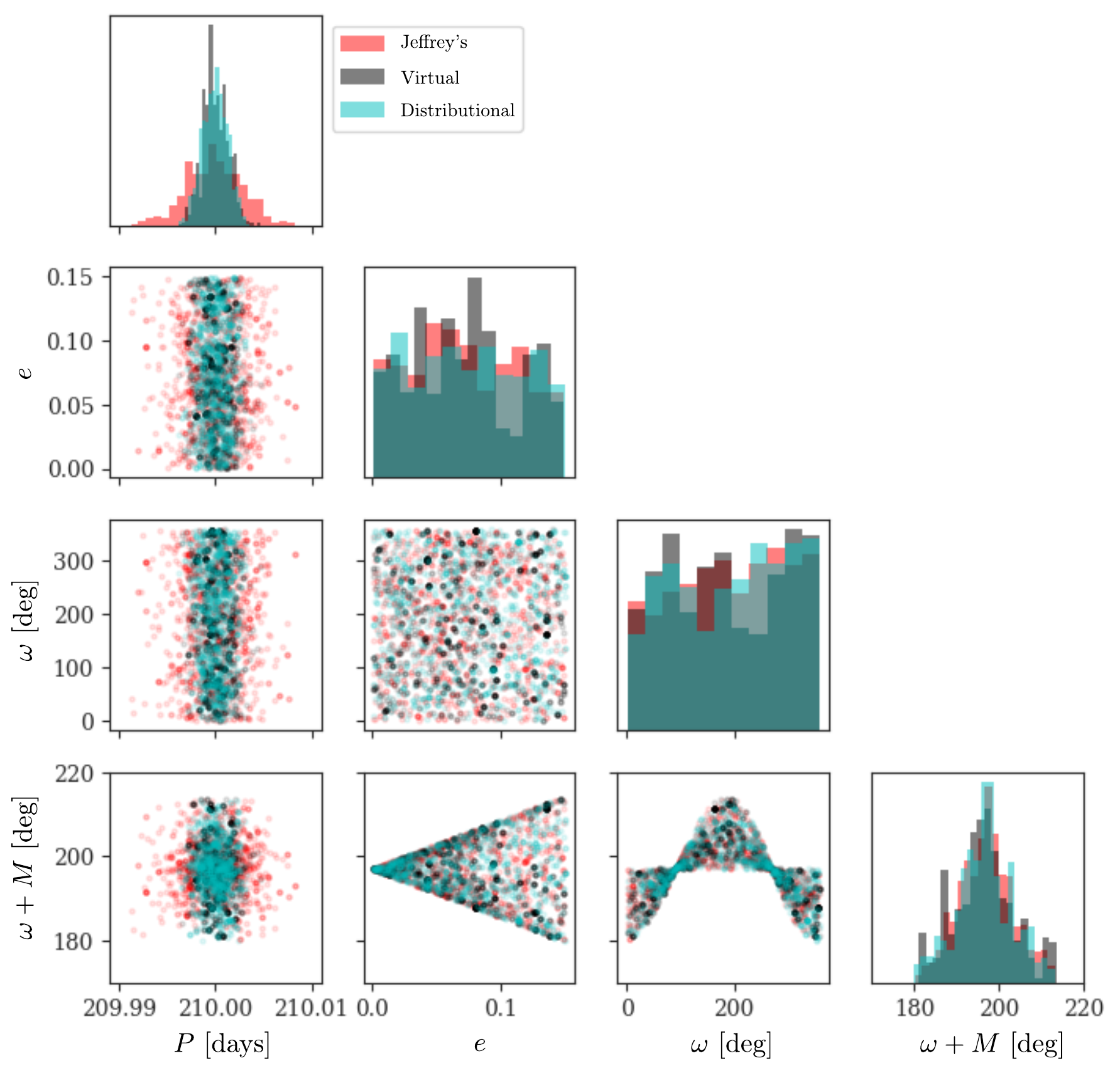}
  \caption{Inferred orbital parameters of an exoplanet around a Kepler star. We
  simulate $7$ transits, $\bit$, of the system, and assume an error on the
  measured transits of $20$ mins ($q(\bit|\zeta)$) while we assume an error on
  our ability to model the transits of $10$ mins, the likelihood. While the
  marginal posterior distributions of $e$, $\omega$ and $\omega+M$ are all in
  agreement, the posterior over $P$ is significantly different when extracted
  using Jeffrey's rule compared to the other two methods.}
  \label{fig:K90}
\end{figure}

The Kepler satellite \citep{Borucki2010} measured the flux from over half a
million stars over 5 years. Dips in the observed flux can occur when a planet
transits in front of the stellar disk, and accurate measurements of the exact
transit times allow us to infer the orbital properties of the planets. However,
the received flux from distant stars varies for other reasons (e.g., stellar
pulsations, telescope temperature) and in principle one should fit a joint
orbit/stellar/telescope model to the observed flux to infer orbital parameters.
However, it is common \citep[e.g.,][]{liang2021kepler90} to extract this
information in two phases: first to fit a model of the star and to extract from
this the \emph{transit times} and second to extract orbital parameters from
these transit times. Thus the measurements of transit times constitute uncertain
evidence, in that they are provided as estimated times with an associated error.
In the case of a single planet, it is only possible to infer the orbital period,
$P$, and the anomaly angle $\omega+M$, while the other (planar) orbital
parameters, eccentricity, $e$, and periapsis argument, $\omega$, remain
marginally unconstrained (but not in correlation with $P$ and $\omega+M$). We
simulate data, based on Kepler-90g, with $P = 210$ days, $e = 0.05$, $\omega =
100$ deg and $\omega+M = 198$ deg using \textsc{ttvfast} \citep{Deck2014} and
approximate posterior distributions using amorized inference within
\textsc{pyprob} \citep{le2017inference, baydin2019efficient}. The prior over $P$
is taken to be normal with $210\pm1$ days.  The prior over eccentricity is taken
to be uniform between $0$ and $0.15$. The angular variables have uniform priors
between $0$ and $360$ deg. In Fig.~\ref{fig:K90} we provide additional
experimental details and show the 1D and 2D marginal posterior distributions
over orbital parameters given the three different approaches to uncertain
evidence. We note that while the marginal posterior distributions of $e$,
$\omega$ and $\omega+M$ are all in agreement, the posterior over $P$ is
significantly different when extracted using Jeffrey's rule compared to when
using the other two methods.

\section{Related Work}\label{sec:extens-altern-appr}

Of important related work is that of~\cite{valtorta2002soft}, who propose an
approach for dealing with uncertain evidence which is in some way an extension
to Jeffrey's rule. Their algorithm, the \textit{soft evidential update method},
is tailored for Bayesian networks (BN) and they incorporate uncertain evidence
by extending the BN with evidence nodes for each new piece of uncertain
evidence. Their approach updates the prior BN (prior to receiving uncertain
evidence), denoted $M_{P}$, by solving for a new ``updated'' BN, $M_{U}$. The
resulting $M_{U}$ minimizes the Kullback-Leibler divergence between $M_{P}$ and
$M_{U}$ under the constraint that the marginal distribution of $M_{U}$ of each
uncertain evidence variable must equal the given distributions. Given a single
piece of uncertain evidence their update method reduces to Jeffrey's rule.
Another approach is that of~\citet{doi:10.1080/03610926.2020.1838545}, which is
similar to, and discussed by~\citet{tolpin2021probabilistic}. The difference of
this approach compared to distributional evidence lies in the definition of the
likelihood $p(\biy\sim D_q|\bix)$, for
which~\citet{doi:10.1080/03610926.2020.1838545} proposes $p(\biy \sim D_q|\bix)
\propto \EE_{q(\biy)}\left[ p(\biy|\bix) \right]$. However, as discussed
by~\citet{tolpin2021probabilistic}, this definition lacks many (what they deem)
desired properties associated with distributional evidence,
\cref{eq:distributional-evidence}.

\section{Conclusions}

We have considered the problem of Bayesian inference when given uncertain
evidence and the importance of its proper interpretation. This involved
discussing and provided new insights into three different approaches in dealing
with uncertain evidence: Jeffrey's rule, virtual evidence, and distributional
evidence. Particularly, this lead to the definition of four types of commonly
encountered uncertain evidence. We have discussed compatibility between a given
probabilistic model and uncertain evidence as defined in terms of consistency.
We have demonstrated in three different experiments how misinterpretations of
the type of uncertain evidence may lead to different inference results. This
illustrates the importance of carefully making the proper interpretation of
uncertain evidence on a case-by-case basis.

\section*{Acknowledgements}

We acknowledge the support of the Natural Sciences and Engineering Research
Council of Canada (NSERC), the Canada CIFAR AI Chairs Program, and the Intel
Parallel Computing Centers program. Additional support was provided by UBC's
Composites Research Network (CRN), Data Science Institute (DSI), and Lawrence
Berkley National Lab (under subcontract 7623401). This research was enabled in
part by technical support and computational resources provided by WestGrid
(www.westgrid.ca), Compute Canada (www.computecanada.ca), and Advanced Research
Computing at the University of British Columbia (arc.ubc.ca).

\bibliography{bibtex.bib}
\bibliographystyle{icml2023}

\newpage
\appendix
\onecolumn

\section{Commutativity of Jeffrey's Rule}\label{app:comm-jeff}

It is well known \citep{diaconis1982updating} that Jeffrey's rule does not
\textit{generally} commute with respect to different pieces of uncertain
evidence, $\epsilon_A, \epsilon_B$. That is, applying Jeffrey's rule first with
respect to $\epsilon_A$ and then subsequently with respect to $\epsilon_B$ is
\textit{not} necessarily equal to applying Jeffrey's rule in the reverse order.
This is easily seen with the following example: Let $\epsilon_A$ and
$\epsilon_B$ carry contradictory information about the same variable $\biy$. For
each piece of uncertain evidence, consider the associated auxiliary variable
$\zeta_A$ and $\zeta_B$ and the densities $q(\biy|\zeta_A)$ and
$q(\biy|\zeta_B)$. Then from Jeffrey's rule we have the updated distribution of
the latent variable $\bix$:
\[
  p(\bix|\zeta_A, \zeta_B) & = \EE_{q(\biy|\zeta_B)}[p(\bix|\biy,\zeta_A)] = \EE_{q(\biy|\zeta_B)}[p(\bix|\biy)] = p(\bix|\zeta_B) \\
  p(\bix|\zeta_B, \zeta_A) & = \EE_{q(\biy|\zeta_A)}[p(\bix|\biy,\zeta_B)] = \EE_{q(\biy|\zeta_A)}[p(\bix|\biy)] = p(\bix|\zeta_A),
\]
where we use $p(\cdot|\zeta_1, \zeta_2)$ as an overloaded denotation for
applying Jeffrey's rule first with respect to $\zeta_1$ and subsequently with
respect to $\zeta_2$. In this example, we see that the second piece of uncertain
evidence dominates and ``overwrites'' or ``forgets'' the first. This illustrates
that if two pieces of ``incompatible'' uncertain evidence are given, care must
be taken when using Jeffrey's rule. We leave the topic of addressing
commutativity of Jeffrey's rule for future discussion, but we briefly mention
that a potential remedy could be to define a mixture of $q(\biy|\zeta_A)$ and
$q(\biy|\zeta_B)$, which would require incorporating $\epsilon_A$ and
$\epsilon_B$ jointly rather than sequentially.

As a final note, we point out the likelihood-bases approaches to uncertain
evidence, such as virtual evidence, does commute with respect to multiple pieces
of uncertain evidence. In particular, given two incompatible pieces of uncertain
evidence and associated auxiliary variables $\zeta_A$ and $\zeta_B$, the joint
density would assign zero probability on that event,
$p(\zeta_A,\zeta_B,\biy,\bix)=0$, which in turn may indicate a misspecification
of the model.

\section{Proofs}\label{app:proofs}

\bprfof{\cref{eq:pearls-discrete}}\label{app:proof-virtual} Consider the
assumptions in \cref{def:virtual-evidence} and let $\biy\in\left\{ \biy_{k}
\right\}_{k=1}^{K}$ be discrete. From \cref{eq:lik-ratios} it follows
that $p(\zeta|\biy_{k})=c \lambda_{k}$ with $k=1,\dots,K$ for some
$c\in\posReals$. This leads to,
\[
  p(\bix|\zeta)
  &=\frac{\sum_{k=1}^{K}p(\bix,\biy_{k},\zeta)}{p(\zeta)}=\frac{\sum_{k=1}^{K}p(\zeta|\biy_{k})p(\bix,\biy_{k})}{\sum_{j=1}^{K}p(\zeta,\biy_{j})}
  \\
  &=\frac{\sum_{k=1}^{K}p(\zeta|\biy_{k})p(\bix,\biy_{k})}{\sum_{j=1}^{K}p(\zeta|\biy_{j})p(\biy_{j})}=\frac{\sum_{k=1}^{K}c\lambda_{k}p(\bix,\biy_{k})}{\sum_{j=1}^{K}c\lambda_{j}p(\biy_{j})}
  \\
  &=\frac{c}{c}\frac{\sum_{k=1}^{K}\lambda_{k}p(\bix,\biy_{k})}{\sum_{j=1}^{K}\lambda_{j}p(\biy_{j})}=\frac{\sum_{k=1}^{K}\lambda_{k}p(\bix,\biy_{k})}{\sum_{j=1}^{K}\lambda_{j}p(\biy_{j})}
\]
\eprfof

\subsection{Proofs for \cref{thm:consistency-jeffrey}}
\label{app:proofs-consistent}

\bprfof{\cref{thm:consistency-jeffrey} (1)}\label{proof:consistency-jeffrey-2}

(Necessary) Given $p(\biy,\bix)$ and uncertain evidence $q(\biy|\zeta)$ we need
to show that, if the approach of Jeffrey's rule is consistent, then
\cref{thm:consistency-jeffrey} (1) is true. Consistency requires that there
exists a joint $p(\zeta,\biy,\bix)=p(\zeta|\biy,\bix)p(\biy,\bix)$ containing
$q(\biy|\zeta)$. This implies finding a $p(\zeta|\biy,\bix)$ such that for all
$\zeta$ and $\biy$:
\begin{align}
  q(\biy|\zeta) & =\frac{\EE_{p(\bix)}\left[
      p(\zeta|\biy,\bix)p(\biy|\bix) \right]}{\EE_{p(\biy)} \left[
      \EE_{p(\bix)}\left[p(\zeta|\biy,\bix)p(\biy|\bix) \right]\right]}
  \\
                 & = \frac{p(\zeta|\biy)p(\biy)}{\EE_{p(\biy)} \left[
      p(\zeta|\biy)
      \right]},
\end{align}
where
$p(\zeta|\biy)=\int p(\zeta|\biy,\bix)p(\biy|\bix)p(\bix)/p(\biy)\dd\bix$. 
That is, if no
such $p(\zeta|\biy)$ exists satisfying \cref{thm:consistency-jeffrey} (1) then
the approach of Jeffrey's rule cannot be consistent.

(Sufficient) Assume there exists $p(\zeta|\biy)$ satisfying
\cref{thm:consistency-jeffrey} (1). Define
$p(\zeta,\biy,\bix)=p(\zeta|\biy)p(\biy|\bix)p(\bix)$ from which it immediately
follows that $p(\biy|\zeta)=q(\biy|\zeta)$. Further, using
d-separation~\citep{pearl1988probabilistic}, it follows that defining
$p(\zeta,\biy,\bix)$ in this way ensures that $p(\bix|\biy,\zeta)=p(\bix|\biy)$
i.e., it satisfies \cref{thm:consistency-jeffrey} (1). This proves Jeffrey's
rule is consistent, such that:
\begin{align}
  p(\bix|\zeta) & =\EE_{p(\biy|\zeta)}\left[ p(\bix|\biy,\zeta) \right] \\
                 & = \EE_{q(\biy|\zeta)}\left[ p(\bix|\biy) \right].
\end{align}
\eprfof

\bprfof{\cref{thm:consistency-jeffrey} (2)} \label{proof:consistency-jeffrey-3}

\begin{figure}[]
  \centering
  \includegraphics{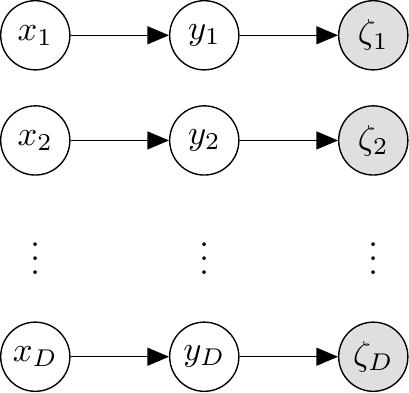}
  \caption{\label{fig:jeffrey-model-independence} }
\end{figure}

Let each $\left\{ y_{i} \right\}_{i=1}^{D}$ be conditionally independent given
$\zeta$ such that $q(\biy|\zeta)=\prod_{i=1}^{D}q(y_{i}|\zeta)$. Further, assume
Jeffrey's rule is consistent such that there exists a joint model
$p(\zeta,\bix)$ where $p(\biy|\zeta)=q(\biy|\zeta)$. Then it follows, via
d-separation, that if and only if all paths between each $\left\{ y_{i}
\right\}_{i=1}^{D}$ are conditionally blocked can they be conditionally
independent given $\zeta$. This implies that no two or more $y_{i}$ can share
the same auxiliary variable or latent variable or depend on each other.
\cref{fig:jeffrey-model-independence} shows the only possible graphical model
satisfying this constraint, which leads to:
\begin{equation}
  p(\biy|\zeta)=\prod_{i=1}^{D}p(y_{i}|\zeta_{i}) \Rightarrow
  p(\biy|\bix)=\prod_{i=1}^{D}p(y_{i}|x_{i}).
\end{equation}
\eprfof

\bprfof{\cref{thm:consistency-jeffrey} (3)}

Assume Jeffrey's rule is consistent such that $q(\biy|\zeta)=p(\biy|\zeta)$
which implies $p(\zeta)= \EE_{p(\biy)}\left[ p(\zeta|\biy) \right]$. From the
law of total variance we have:
\begin{equation} \label{eq:law-total-variance}
  \cov\left[ \biy \right] = \EE\left[ \cov\left[ \biy \vert \zeta
      \right]\right] + \cov \left[ \EE\left[ \biy \vert \zeta \right] \right],
\end{equation}
with the right-hand side being a sum of two positive semi-definite matrices.
Since for two positive semi-definite matrices $A$ and $B$ it holds that
$\det(A+B)\geq \det(A) + \det(B)$ \citep{paksoy2014inequalities}, and as
$\det(A), \det(B) \geq 0$ this leads to:
\begin{equation}
  \cov\left[ \biy \right] \succeq \EE\left[ \cov\left[ \biy |\zeta
      \right]\biy \right].
\end{equation}
Further, we have that the elements in the diagonal of the left-hand side of
\cref{eq:law-total-variance} are the variances of $\biy$ with respect to
$p(\biy)$ and therefore:
\begin{align}
  \var\left[ y_{i} \right] & = \EE \left[ \var\left[
      y_{i} \vert \zeta	  \right]\right] + \var \left[
    \EE\left[
      y_{i} \vert \zeta \right] \right]
  \label{eq:law-total-var-scalar}
  \\
                                          & \geq \EE\left[ \var\left[
      y_{i} \vert \zeta	\right]\right],
  ~ \mathrm{as}~\var \left[ \EE\left[ y_{i} \vert \zeta
      \right]\right]\geq 0.
\end{align}
Finally we prove that:
\begin{equation}
  \var\left[ y_{i} \right] = \EE\left[ \var\left[ y_{i} \vert \zeta	     \right]\right] \Leftrightarrow \EE\left[ y_{i} \vert \zeta \right] = \mu.
\end{equation}
We first prove ``$\Rightarrow$'':
\begin{equation}
  \var\left[ y_{i} \right] = \EE\left[ \var\left[
      y_{i} \vert \zeta	     \right]\right] \Rightarrow
  \var\left[
    \EE\left[ y_{i} \vert \zeta \right] \right] = 0.
\end{equation}
As $\var \left[ x \right]=\EE \left[ \left( x- \EE \left[ x \right] \right)^{2}
\right]$ is an expectation of a non-negative variable, it follows that
$\var\left[ x \right]=0$ if and only if $x$ is constant. Therefore, we have
that:
\begin{equation}\label{eq:const-mean-zero-var}
  \var\left[ \EE\left[ y_{i} \vert \zeta \right] \right] = 0
  \Leftrightarrow \EE\left[ y_{i} \vert \zeta \right] = \mu,
\end{equation}
where $\mu$ is some constant.

Next we prove ``$\Leftarrow$'' which follows trivially by combining
\cref{eq:const-mean-zero-var} with \cref{eq:law-total-var-scalar}:
\begin{equation}
  \var\left[ \EE\left[ y_{i} \vert \zeta \right] \right] = 0
  \Rightarrow \var\left[ y_{i} \right] = \EE\left[ \var\left[
      y_{i} \vert \zeta	\right]\right].
\end{equation}
From this we can conclude that:
\begin{equation}
  \var\left[ y_{i} \right] = \EE\left[ \var\left[ y_{i} \vert \zeta	     \right]\right] \Leftrightarrow \EE\left[ y_{i} \vert \zeta \right] = \mu,
\end{equation}
thereby concluding the proof.
\eprfof

\section{Other Derivations}

\subsection{Distributional Evidence and Normal Distributions}\label{app:distributional-evi-form} 

Consider the densities $p(y|x)=\distNorm(\mu_{y|x}|\sigma_{y|x}^2)$ and
$q(y)=\distNorm(\mu_q|\sigma_q^2)$ and the distributional evidence likelihood,
\cref{eq:distributional-evidence}:
\[
  \ln p(y\sim D_q |x) \propto \EE_q\left[\ln p(y|x)\right]
    &= -\frac{1}{2\sigma_{y|x}^2}\EE_q\left[\left(y - \mu_{y|x} \right)^2\right] - \ln\left(\sqrt{2\pi}\sigma_{y|x}\right) \\
    & \tikzmark{a}= -\frac{1}{2\sigma_{y|x}^2}\left[\EE_q[y^2] - 2\mu_q\mu_{y|x}^2 + \mu_{y|x}^2\right]  - \ln\left(\sqrt{2\pi}\sigma_{y|x}\right) \\
    & \tikzmark{b}= -\frac{1}{2\sigma_{y|x}^2}\left[\mu_q^2 - 2\mu_q\mu_{y|x}^2 + \mu_{y|x}^2\right] - \ln\left(\sqrt{2\pi}\sigma_{y|x}\right) \\
    & =  -\frac{1}{2\sigma_{y|x}^2}\left(\mu_q - \mu_{y|x}\right)^2 - \ln\left(\sqrt{2\pi}\sigma_{y|x}\right).\label{eq:sto-loglik-gaus-propto}
    \begin{tikzpicture}[overlay, remember picture, distance=0.4cm]
      \path(a.center) edge [->, shorten >=2pt, shorten <=2pt, bend right=50] node
        [midway,align=center, anchor=east] {\tiny $\EE_q[y^2]=\sigma_q^2 + \mu_y^2$} (b.center);
  \end{tikzpicture}
\]
Assuming the distribution $D_q$ is implicitly defined via its mean $\mu_q$, such that
$p(y\sim D_q)$ normalizes with respect to $\mu_q$, we identify from
\cref{eq:sto-loglik-gaus-propto} $p(y\sim D_q|x)$ to be a Gaussian with mean
$\mu_{y|x}$ and variance $\sigma_{y|x}^2$. From this we see that distributional
evidence in this special case leads to the \textit{same} likelihood as the one
in the base model, $p(y=\mu_q|x)$. That is, $p(y\sim D_q|x)=p(y=\mu_q|x)$.
Further, we find that the normalization constant
\[
  Z(x) = \int \exp \EE_q\left[\ln p(y|x)\right] \mathrm{d}\mu_q
  =\frac{\sqrt{2\pi}\sigma_{y|x}}{\sqrt{2\pi}\sigma_{y|x}}=1,
\]
is independent of $x$. This leads to the distributional evidence adjusted prior
$p_\mathrm{a}(x)\propto p(x)Z(x)=p(x)$ being equal to the non-adjusted prior $p(x)$
\[
  p_\mathrm{a}(x)= \frac{p(x)}{\int p(x)\mathrm{d}x} = p(x).
\]
In this case, $p(x|y\sim D_q)$ always takes the form $p(x|y\sim D_q)\propto q(y
\sim D_q | x)p(x)$. Further, we can generally say that if
$p(y|x)=\distNorm(\mu_{y|x}|\sigma^2)$ and $q(y)=\distNorm(\mu_q|\sigma_q^2)$
then posterior inference using distributional evidence given uncertain evidence
reduces to exact evidence in the base model $p(x,y)$ given \textit{exact}
evidence, $p(x|y=\mu_q)$. These results generalizes to the multivariate case
where the likelihood in the base model and the distributional evidence density
are defined on an observable vector $\biy=\left(y_1,\dots,y_K\right)$ and each
$y_k$ is \iid such that $p(\biy|x)=\prod_{k=1}^K p(y_k|x)$ and
$q(\biy|x)=\prod_{k=1}^K q(y_k|x)$.

\end{document}